\documentclass[reqno,12pt]{amsart}

\usepackage{url}
\usepackage{bm}
\usepackage{amssymb}
\usepackage[symbol]{footmisc}

\makeatletter
\def\blfootnote{\gdef\@thefnmark{}\@footnotetext}
\makeatother

\newcommand\bi{\bm i}
\newcommand\bj{\bm j}
\newcommand\bk{\bm k}

\newcommand\starop[1]{\mathsf S(#1)}

\newcommand\setdualquat{\mathfrak d \mathfrak h}
\newcommand\setunitdualquat{\mathbb S\mathbb D\mathbb H}
\newcommand\setvectordualquat{\mathfrak{s}\mathfrak d \mathfrak h}

\DeclareMathOperator\realpart{Re}
\DeclareMathOperator\imagpart{Im}
\newcommand{\liediff}{\mathbin{\triangle}}
\newcommand{\liederiv}{\mathcal L}

\begin{document}

\title{An introduction to using dual quaternions to study kinematics}

\author{Stephen Montgomery-Smith}
\address{Department of Mathematics, University of Missouri, Columbia, MO 65211.}
\email{stephen@missouri.edu}
\urladdr{https://stephenmontgomerysmith.github.io}

\author{Cecil Shy}
\address{Johnson Space Center, 2101 E.~NASA Pkwy, Houston, TX 77058.}
\email{cecil.shy-1@nasa.gov}

\date{July 22, 2023}

\begin{abstract}
We explain the use of dual quaternions to represent poses, twists, and wrenches. \end{abstract}

\maketitle


\section{Introduction}

We assume that the reader is familiar with the notion of \emph{pose} and \emph{rigid motion}, and the use of quaternions to represent rotations.  We use the terms pose and rigid motion interchangeably, since a pose can be considered as a rigid motion relative to a fixed reference frame, and the mathematical operations to manipulate them are identical.  Thus we also use the terms rotation and orientation interchangeably, and similarly with translation and position.  We refer the reader to \cite{bottema-et-al,gallardo-alvarado,selig-book,spong-et-al}.

The use of dual quaternions to represent poses is well established,: \cite{adorno,agrawal,clifford,dooley-et-al,han-et-al,kavan-et-al,kavan-et-al-2,kenwright,kussaba-et-al,perez-et-al,schilling1,schilling2,silva-et-al,wang-et-al,yang-et-al}.

Let us start with a general discussion of why people might want to represent rotations by unit quaternions instead of three by three orthogonal matrices with determinant $1$.  One reason given is that quaternions can be represented with only four numbers, whereas the matrix representation requires nine numbers.  Another representation commonly used are Euler angles, and these only require three numbers.  However, with Euler angles, the formula for composition of rotations is difficult to calculate.  Also, Euler angles suffer from gimbal lock, in which if the rotation takes certain values, the choice of Euler angles isn't unique, and discontinuities and singularities arise.  Quaternions suffer from no such problems.

But in modern times, computer memory is very cost effective, and communication is fast, thus the aforementioned advantages might seem trivial at first.  This begs the question: what other advantages do quaternions offer over matrices?  Quaternions represent a quantity with three degrees of freedom by a four dimensional vector, whereas matrices use a nine dimensional vector.  Suppose one obtains an approximation to a matrix or quaternion?  For example, this would be common when using Newton-Raphson to find a rotation satisfying some formula, or to interpolate between rotations.  So then one has to find a projection which takes the vector to an admissible vector, that is, a vector that actually represents a rotation.  And projecting from a nine dimensional quantity to a three dimensional quantity is going to be much harder than the similarly projection from the four dimensional vector.  (The projection from quaternions to unit quaternions is called the \emph{normalization}, and is defined in Section\ref{sec motivation}.  A suggestion of what such a projection would be for $(3 \times 3)$ matrices is given at the end of Section~\ref{sec final note}.)

For representing poses, the usual representation is by a three by three orthogonal matrix with determinant $1$ for the rotation, and a three dimensional vector for the translation, twelve dimensional in all.  This has a natural representation as a four by four matrix, as we explain Section~\ref{sec twists}.  A great advantage of this representation is the relationship between pose and twist:
\begin{equation}
\label{pose twist}
\frac d{dt} (\text{pose}) = (\text{pose}) \times (\text{twist}).
\end{equation}
where the multiplication is matrix multiplication, and a twist is a combination of angular velocity and translational velocity measured with respect to the moving frame.

Another natural representation used a lot is to use a quaternion and a three dimensional vector.  But this new representation has problems in that it is not clear how the above equation would work in this situation.  However, if one introduces the notion of the dual quaternion, which is an eight dimensional vector, equation~\eqref{pose twist} is still valid if one uses dual quaternion multiplication.

A challenge with dual quaternions is that they are abstract, which can make them harder to understand.  Mathematicians find value behind abstract computational ideas, but Engineers usually find the end result, `black box approach,' more useful.

A useful approach to developing intuition for dual quaternions is to understand the rules with their structure, and the mathematical operations needed to manipulate them, (that is, normalization, basis vector multiplication, pose differentiation).  Once one becomes comfortable with their form and function, the advantage of duel quaternions over other methods becomes more apparent.  For example, one realizes that computing the projection of an eight dimensional vector to the six degrees of freedom of a pose, is also quite simple.  (This projection is also called the normalization, and is defined in Section~\ref{sec poses}.)

This makes it very useful if one is using the Newton-Raphson Method to find a pose satisfying certain properties.  We show how the dual quaternion representation efficiently solves the forward kinematics problem.

Finally we consider the dynamics of poses, that is, the equations of motion of a rigid spinning body.  In any representation this can be quite difficult, because of both centripetal and precessional effects.  But from a theoretical perspective, using the dual quaternion representation is quite slick.

In summary, while the usual representation using Euler angles and translations is conceptually easy to understand, performing calculations can be difficult.  Whereas with dual quaternions, while the are initially conceptually hard to understand, when it comes to calculations they are much easier to work with.

\section{Motivation}
\label{sec motivation}

A \emph{pose}, or \emph{rigid motion}, is a rotation followed by a translation.  (Semantically, a pose is the position of the end effector of the robot, and thus different from a rigid motion, but mathematically they are the same if we consider a pose to be a rigid motion with respect to the stationary reference frame.  But see \cite{chirikjian-et-al} for a different point of view.)  The rotation is represented by a $(3 \times 3)$ orthogonal matrix $\mathsf R$ with determinant $1$.  A translation $\bm t$ is a vector in $\mathbb R^3$.  These represent the pose $(\mathsf R, \bm t)$:
\begin{equation}
\label{pose action}
\bm r \mapsto \mathsf R \bm r + \bm t .
\end{equation}
Composition of poses is written from right to left.  Thus
\begin{equation}
(\mathsf R_1, \bm t_1) \circ (\mathsf R_2, \bm t_2) = (\mathsf R_1 \mathsf R_2, \bm t_1 + \mathsf R_1 \bm t_2).
\end{equation}
One way to represent a pose is using a four by four matrix
\begin{equation}
\label{pose-matrix}
\begin{bmatrix} \mathsf R & \bm t \\ 0 & 1 \end{bmatrix}
 = \begin{bmatrix} \mathsf R_{11} & \mathsf R_{12} & \mathsf R_{13} & t_1 \\
                   \mathsf R_{21} & \mathsf R_{22} & \mathsf R_{23} & t_2 \\
                   \mathsf R_{31} & \mathsf R_{32} & \mathsf R_{33} & t_3 \\
                   0 & 0 & 0 & 1 \end{bmatrix} 
\end{equation}
Then composition of poses may be computed simply by multiplying matrices of the form~\eqref{pose-matrix}.  In the literature, the Lie group of rotations is often denoted $SO(3)$.  The Lie group of poses is often denoted $SE(3)$.

The problem with this representation is that twelve numbers are required to represent a pose.  Other problems, which we discuss below, is that of interpolating a sequence of poses, and of computing forward kinematics.

One common way to solve both of these problems is to represent the rotation by a unit quaternion \cite{quaternions1,quaternions2}, which we briefly describe here.  A \emph{quaternion} is a quadruple of real numbers, written as $A = w + x \bi + y \bj + z \bk$, with the algebraic operations $\bi^2 = \bj^2 = \bk^2 = \bi \bj \bk = -1$.  Its \emph{conjugate} is $A^* = w - x \bi - y \bj - y \bk$, its \emph{norm} is $|A| = (w^2+x^2+y^2+z^2)^{1/2} = \sqrt{A A^*} = \sqrt{A^* A}$, its \emph{normalization} is $\widehat A = A/|A|$, its \emph{real part} is $\realpart(A) = w = \tfrac12(A + A^*)$, and its \emph{imaginary part} is $\imagpart(A) = \bi x + \bj y + \bk z = \tfrac12(A - A^*)$.  It is called a \emph{unit} quaternion if $|A| = 1$, a \emph{real} quaternion if $\imagpart(A) = 0$, and a \emph{vector} quaternion if $\realpart(A) = 0$.  Note the multiplicative inverse is given by $A^{-1} = A^*/|A|^2$.

We identify three dimensional vectors with vector quaternions, by identifying $\bi$, $\bj$, and $\bk$ with the three standard unit vectors.  A unit quaternion $Q$ represents the rotation $\bm r \mapsto Q \bm r Q^*$.  A rotation by angle $a$ about an axis $\bm s$, where $|\bm s| = 1$, has two unit quaternion representations: $\pm(\cos(\tfrac12 a) + \bm s \sin(\tfrac12 a))$.  Composition of rotations corresponds to multiplication of unit quaternions.  With some practice, it becomes easier to read a rotation from a quaternion than it does from Euler angles, the standard method of representing rotations.

We can represent quaternions as four dimensional vectors, and give it the inner product
\begin{equation}
A \cdot B = \realpart(A B^*) = \realpart(A^* B) .
\end{equation}

Then a way to represent poses is as a pair $(Q,\bm t)$.  This is the way poses are represented internally in the Robotic Operating System \cite{ros}.  Then the composition rule is
\begin{equation}
\label{pose-ros-compose}
(Q_1, \bm t_1) \circ (Q_2, \bm t_2) = (Q_1 Q_2, \bm t_1 + Q_1 \bm t_2 Q_1^*).
\end{equation}
A difficulty with this is that the composition is not a bilinear operation as it was with the representation~\eqref{pose-matrix}.  (This lack of bilinear representation becomes particularly problematic when representing twists, which we describe below.)  However there is a way to resolve this as follows.  Write $B_1 = \tfrac12 \bm t_1 Q_1$ and $B_2 = \tfrac12 \bm t_2 Q_2$.  Then equation~\eqref{pose-ros-compose} can be rewritten as
\begin{equation}
\begin{aligned}
& (Q_1, 2 B_1 Q_1^{-1}) \circ (Q_2, 2 B_2 Q_2^{-1}) 
\\& = (Q_1 Q_2, 2 B_1 Q_1^{-1}+ 2 Q_1 B_2 Q_2^{-1} Q_1^{-1})
\\& = (Q_1 Q_2, 2 (B_1 Q_2 + Q_1 B_2)(Q_1 Q_2)^{-1}) .
\end{aligned}
\end{equation}
This leads to representing the pose $(Q,\bm t)$ by a unit dual quaternion, which we now describe.

\section{Dual quaternions to represent poses}
\label{sec poses}

A \emph{dual quaternion} is a pair of quaternions, written as $\eta = A + \epsilon B$, with the extra algebraic operation $\epsilon^2 = 0$.

The \emph{conjugate} dual quaternion of $\eta = \eta = A + \epsilon B$ is $\eta^* = A^* + \epsilon B^*$.  Conjugation reverses the order of multiplication:
\begin{equation}
(\eta_1\eta_2)^* = \eta_2^* \eta_1^* .
\end{equation}
There is another conjugation for dual quaternions: $\overline{(A + \epsilon B)} = A - \epsilon B$, but we have no cause to use it in this paper, except in equation~\eqref{defn of pose on 3-vector} below.

A \emph{unit} dual quaternion $\eta = Q + \epsilon B$ is a dual quaternion such that $\eta\eta^* = 1$, equivalently, that $Q$ is a unit quaternion and $B \cdot Q = 0$.  A \emph{vector} dual quaternion $A + \epsilon B$ is a dual quaternion such that both $A$ and $B$ are vector quaternions.

If $\eta = Q + \epsilon B$ is a dual quaternion with $Q \ne 0$, then its multiplicative inverse can be calculated using the formula
\begin{equation}
\eta^{-1} = Q^{-1} - \epsilon Q^{-1} B Q^{-1}.
\end{equation}
If $\eta$ is a unit dual quaternion, then there is a computationally much faster formula:
\begin{equation}
\label{inverse unit}
\eta^{-1} = \eta^*.
\end{equation}

A \emph{dual number} is anything of the form $a + \epsilon b$, where $a$ and $b$ are real numbers.  The \emph{norm} of a dual quaternion $\eta = Q + \epsilon B$ is the dual number defined by the two steps:
\begin{gather}
|\eta|^2 = \eta^* \eta = \eta \eta^* = |Q|^2 + 2 \epsilon (B \cdot Q) ,\\
\label{norm}
|\eta| = \sqrt{|\eta|^2} = |Q| + \epsilon (B \cdot Q) / |Q| .
\end{gather}
The norm preserves multiplication, that is, if $\eta_1$ and $\eta_2$ are two dual quaternions, then
\begin{equation}
\label{norm mult}
|\eta_1 \eta_2| = |\eta_1| |\eta_2| .
\end{equation}

If $\eta = Q + \epsilon B$ is any dual quaternion with $Q \ne 0$, then we define its \emph{normalization} to be the unit dual quaternion
\begin{equation}
\label{normalize}
\widehat \eta = |\eta|^{-1}\eta = \eta|\eta|^{-1}
= Q/|Q| + \epsilon (B/|Q| - (B\cdot Q) Q/|Q|^3) .
\footnote[2]{Note this formula is incorrect in earlier versions of this paper.  The authors are grateful to Brent Koogler for bringing this to our attention.}
\end{equation}
(We remark that the normalization of a dual quaternion is used in the computer graphics industry \cite{kavan-et-al, kavan-et-al-2}.)  While this normalization formula might seem initially quite complicated, after thinking about it one can see that it is the simplest projection that enforces $|Q| = 1$ and $B\cdot Q = 0$.

The normalization also satisfies the following properties.
\begin{itemize}
\item If $\eta$ is a unit dual quaternion, then $\widehat\eta = \eta$.
\item Normalization preserves multiplication, that is, if $\eta_1$ and $\eta_2$ are two dual quaternions, then
\begin{equation}
\label{normalize mult}
\widehat{\eta_1 \eta_2} = \widehat \eta_1 \widehat \eta_2 .
\end{equation}
\end{itemize}
Note that
\begin{equation}
\eta^{-1} = |\eta|^{-2} \eta^*,
\end{equation}
and this could have been the definition of the multiplicative inverse of a dual quaternion, except that this definition is circular.

We set $\setdualquat$ for the set of dual quaternions, $\setunitdualquat$ for the set of unit dual quaternions, and $\setvectordualquat$ for the set of vector dual quaternions.

If we are given a pose represented by $(Q, \bm t)$, then the pose is also represented by the unit dual quaternion
\begin{equation}
\label{pose as dual quaternion}
\eta = Q + \tfrac12 \epsilon \bm t Q.
\end{equation}
As we have shown above, composition of poses corresponds to multiplication of unit dual quaternions.  If $\bm r$ is a 3-vector, and $\bm s$ is the image of $\bm r$ under the action of the pose $\eta = Q + \epsilon B$, then
\begin{equation}
\label{defn of pose on 3-vector}
1 + \epsilon \bm s = \eta (1 + \epsilon \bm r) \overline\eta^* ,
\end{equation}
but generally it is easier to use the formula
\begin{equation}
\label{pose on 3-vector}
\bm s = Q \bm r Q^* + 2 B Q^* = (Q \bm r + 2 B) Q^* .
\end{equation}
The introduction of the factors $2$ and $\frac12$ is unnecessary, but it slightly improves formulas for the twist, as we see below.

It is difficult for a human to look at the second part of a unit dual quaternion, and from that read what the translational part of the pose should be.  So dual quaternions work better for an internal representations of poses rather than human readable representations.

Finally, let us attempt to provide a natural explanation of equation~\eqref{defn of pose on 3-vector}.  First note that if $\eta = Q + \tfrac12 \epsilon \bm t Q$ is the unit dual quaternion representing the pose $(Q, \bm t)$, then one way to recover the translation part is as follows:
\begin{equation}
\eta \overline \eta ^* = (1 + \tfrac12 \epsilon \bm t) Q \overline{((1 + \tfrac12 \epsilon \bm t) Q)}^*= (1 + \tfrac12 \epsilon \bm t)Q Q^* (1 + \tfrac12 \epsilon \bm t) = 1 + \epsilon \bm t .
\end{equation}
If we are given a vector $\bm r$, then this is the translation part of a dual quaternion
\begin{equation}
\nu _ {\bm r} = 1 + \tfrac12 \epsilon \bm r .
\end{equation}
Saying $\bm s$ is the image of $\bm r$ under the action of the pose $(Q, \bm t)$ is the same statement as saying that the translational part of $\eta \nu_{\bm r}$ is $\bm s$, that is
\begin{equation}
\eta \nu_{\bm r} \overline {(\eta \nu_{\bm r})}^* = 1 + \epsilon \bm s .
\end{equation}
Since the left hand side simplifies to
\begin{equation}
\eta \nu_{\bm r} \overline{\nu_{\bm r}}^* \overline \eta^* = \eta \nu_{\bm r} (1 + \epsilon \bm r) \overline \eta^*,
\end{equation}
equation~\eqref{defn of pose on 3-vector} follows.

\section{Kinematics: dual quaternions to represent twists}
\label{sec twists}

The translational velocity $\bm v$ of a translating reference frame is given by $\bm v = \dot{\bm t}$.  The angular velocity $\bm w = (w_1, w_2, w_3)$ of a rotating frame is given by the differential equation
\begin{equation}
\label{twist matrix}
\dot{\mathsf R} = \mathsf R \starop{\bm w},
\end{equation}
or $\starop{\bm w} = \mathsf R^{-1} \dot{\mathsf R}$,
where $\starop{\bm w}$ is the \emph{Hodge star operator} of $\bm w$ \cite{wiki0}:
\begin{equation}
\starop{\bm w} = \begin{bmatrix}  0   & -w_3 &  w_2 \\
                               w_3 &  0   & -w_1 \\
                              -w_2 &  w_1 &  0 \end{bmatrix} .
\end{equation}
(Equation~\eqref{twist matrix} gives the angular velocity with respect to the rotating reference frame.  If one wants the angular velocity with respect to the stationary reference frame, use instead $\dot{\mathsf R} = \starop{\bm w} \mathsf R$.)

Note that
\begin{equation}
\label{w star to w cross}
\starop{\bm w} \bm r = \bm w \times \bm r ,
\end{equation}
where $\times$ denotes the cross product.

A \emph{twist} is the pair of vectors $(\bm w, \bm v)$ that describe the change of pose in the moving reference frame, that is:
\begin{equation}
\label{ode twist}
\frac d{dt} \begin{bmatrix} \mathsf R & \bm t \\ 0 & 1 \end{bmatrix}
= \begin{bmatrix} \mathsf R & \bm t \\ 0 & 1 \end{bmatrix}
\begin{bmatrix} \starop{\bm w} & \bm v \\ 0 & 0 \end{bmatrix}.
\end{equation}
(Note in particular that $\dot{\bm t} = \mathsf R \bm v$, so that a continual application of a twist can result in a motion following a straight line, an arc of a circle, or a spiral.)  The set of angular velocities, and the set of twists are both Lie algebras, often denoted by $\mathfrak{so}(3)$ and $\mathfrak{se}(3)$ in the literature.

The quaternion-translation formulation doesn't have a representation like equation~\eqref{pose twist}:
\begin{equation}
\frac d{dt} (Q, \bm t) = \left(\tfrac12 Q \bm w, Q \bm v Q^* \right).
\end{equation}
But with the unit dual representation, because the composition of poses is given by a bilinear multiplication, it follows that the twist is represented by the dual quaternion $\varphi$ satisfying the differential equation.
\begin{equation}
\label{ode dual quaternion}
\varphi = \eta^{-1} \dot \eta, \quad\text{or}\quad\dot \eta = \eta \varphi .
\end{equation}
It will be shown in Section~\ref{sec proofs} (see also \cite{han-et-al,wang-et-al}) that $\varphi$ is the vector dual quaternion
\begin{equation}
\label{twist as dual quaternion}
\varphi = \tfrac12 \bm w + \tfrac12 \epsilon \bm v .
\end{equation}
Note that if $\eta$ is a unit dual quaternion that is a function of time $t$, then $\eta^{-1} \dot \eta$ is always a vector dual quaternion.  (Note that $\dot\eta\eta^{-1}$ is a vector dual quaternion equal to the twist relative to the fixed frame of reference.  Note also that the factor $\tfrac12$ attached to $\bm v$ in equation~\eqref{twist as dual quaternion} is a consequence of the introduction of the factor $\tfrac12$ in equation~\eqref{pose as dual quaternion}, and thus is essentially arbitrary.)

\section{Dual quaternions to represent wrenches}

Let the pose $\eta$ represent the reference frame that moves with the end effector.  It is not necessary (although it can simplify things) that the center of mass of the end effector coincides with the origin of the moving frame.

The \emph{wrench dual quaternion} is defined to be
\begin{equation}
\label{wrench as quaternion}
\tau = 2 \bm q + 2 \epsilon \bm p,
\end{equation}
where $\bm q$ and $\bm p$ are the torque and force, respectively, applied to the end effector at the origin of the moving frame, measured with respect to the moving frame.

If $\bm r_0$ is the center of mass of the end effector in the moving frame, then the twist about the center of mass is given by
\begin{equation}
\label{twist correction}
\varphi_0 = 
\varphi + \tfrac12\epsilon\bm w \times \bm r_0,
\end{equation}
where $\varphi = \eta^{-1} \dot \eta$, and the wrench applied about the center of mass is
\begin{equation}
\label{torque correction}
\tau_0 = 
\tau + 2 \bm p \times \bm r_0.
\end{equation}

The reason for introducing the factor $2$ in definition~\eqref{wrench as quaternion} is so that the rate of change of work done to the end effector is given by
\begin{equation}
\label{dot h tau varphi}
\frac{d}{dt} \text{(work done)} = \tau \cdot \varphi = \tau_0 \cdot \varphi_0 .
\end{equation}

See \cite{ball} for the origins of the term twist and wrench as pairs of 3-vectors, which are examples of \emph{screws}.

\section{Interpolation of poses}

Suppose that we are given a sequence of times and rotations, $t_0 < t_1 < \dots < t_n$ and $\mathsf R_0, \mathsf R_1, \dots, \mathsf R_n$.  We would like to find a function $\mathsf F$ from $[t_1,t_n]$ to rotations such that the third derivative of $\mathsf F$ is bounded, and $\mathsf F(t_k) = \mathsf R_k$.  If the quantities $\mathsf R_0, \mathsf R_1, \dots, \mathsf R_n$ were merely vectors, we could use cubic splines.  But if we perform a similar calculation on the matrices, we cannot guarantee that the matrix $\mathsf F(t)$ is a rotation matrix.

So what we want is to find a map that projects general matrices onto rotation matrices.  But it is not obvious how to find such a map that is computationally efficient.  It is here that using quaternions really has great advantages over matrices, because an obvious map from general quaternions to unit quaternions is to normalize.

Let $Q_0,Q_1,\dots,Q_n$ be the quaternion representation of $R_0, R_1, \dots, R_n$ respectively.  Let $G(t)$ be the cubic spline interpolation such that $G(t_k) = Q_k$.  Then
\begin{equation}
F(t) = \frac1{|G(t)|} G(t)
\end{equation}
provides a computationally fast, low jerk, method of interpolating the rotations.

This can be extended to interpolate poses.  Given times $t_0<t_1<\dots <t_n$ and unit dual quaternions $\eta_0, \eta_1, \dots, \eta_n$, first interpolate to get a function $\gamma$ such that $\gamma(t_k) = \alpha_k$, and then normalize it to get a function $\eta(t) = \widehat{\gamma(t)}$.  This gives a computationally fast, low jerk, method for interpolating poses.

It is perhaps better to interpolate the rotation quaternions and the translations separately.  This is because the center of gravity typically travels in a straight line rather than the arc of a curve, but naive interpolation of dual quaternions follows arcs of curves if the angular rotation is non-zero.  (Interpolation of dual quaternions is more appropriate for `skinning' \cite{kavan-et-al,kavan-et-al-2}.)

If one wants to find a function that passes through these points, and has low jerk even at the end points $t_0$ and $t_n$, this can be done by adding a `ramp up' and `ramp down' at the ends of the trajectory.  Create a free cubic spline $\gamma(t)$ on $\tfrac12(t_0+t_1)<t_1\dots<t_{n-1}<\tfrac12(t_{n-1}+t_n)$ and $\eta_0, \eta_1, \dots, \eta_n$.  Then define a function $s:\mathbb R \to [\tfrac12(t_0+t_1),\tfrac12(t_{n-1}+t_n)]$
\begin{equation}
s(t) = 
\begin{cases}
\tfrac12(t_0+t_1) & \text{if $t < t_0$} \\
t_0 + (t_1-t_0) f \left(\frac{t-t_0}{t_1-t_0}\right) & \text{if $t_0 \le t < t_1$} \\
t & \text{if $t_1 \le t < t_{n-1}$} \\
t_n - (t_n-t_{n-1}) f \left(\frac{t_n-t}{t_n-t_{n-1}}\right) & \text{if $t_{n-1} \le t < t_n$} \\
\tfrac12(t_{n-1}+t_n) & \text{if $t \ge t_n$}
\end{cases}
\end{equation}
where
\begin{equation}
f(t) = \tfrac12 + \tfrac12 (2-t) t^3 .
\end{equation}
Since $f(0) = \tfrac12$, $f'(0) = f''(0) = f''(1) = 0$, and $f(1) = f'(1) = 1$, it follows that $s(t)$ has bounded third derivative, $s(t_k) = t_k$ for $1 \le k \le n-1$, $s(t) = \tfrac12(t_0+t_1)$ for $t \le t_0$, and $s(t) = \tfrac12(t_{n-1}+t_n)$ for $t \ge t_n$.
Thus
\begin{equation}
\eta(t) = \widehat{\gamma(s(t))}
\end{equation}
provides a low jerk function, even at the end points, that passes through the appropriate points.  it is important that $t_1-t_0$ and $t_n-t_{n-1}$ be large enough to allow the trajectory to `ramp up' and `ramp down' smoothly.

\section{Perturbations of poses}

Suppose one has a pose $\eta = \eta(t)$ that varies in time, and a fixed time $t_0$.  Then we would like to find a good representation of the perturbation of $\eta(t)$ from a reference pose $\eta_r = \eta(t_0)$ when $t$ is close to $t_0$.  One way to do this is to choose a new reference frame in which $\eta_r = 1$, the identity pose, or rather, to think about how $\eta_r ^{-1} \eta$ behaves.

It can be seen that for $t$ close to $t_0$ that there is a vector dual quaternion $\theta$ such that
\begin{equation}
\eta = \eta_r \widehat{(1 + \theta)} \approx \eta_r (1 + \theta) .
\end{equation}
Thus
\begin{equation}
\label{proto approx lie diff}
\theta \approx \eta_r^{-1} \eta - 1 = \eta_r^* \eta - 1
\end{equation}
gives a great way to measure the perturbation.

We prefer instead to use what we shall call the \emph{Lie difference}, which we denote using $\liediff$, which is close to \eqref{proto approx lie diff}, but is always a vector dual quaternion:
\begin{equation}
\theta \approx \eta\liediff\eta_r = \tfrac12 (\eta_r^* \eta - \eta^* \eta_r) = \imagpart(\eta_r^* \eta).
\end{equation}
Note that if $\theta_1$ and $\theta_2$ are vector dual quaternions, and $\eta$ is an invertible dual quaternion, then
\begin{equation}
\label{approx approx lie diff}
(\eta(1+\theta_1)) \liediff (\eta(1+\theta_2)) = \theta_1 - \theta_2 + O(|\theta_1|^2 + |\theta_2|^2) .
\end{equation}
This gives a great way to map perturbations of non-linear poses to linear twists.  These can be very effective in control theory.

\section{Spherical linear interpolation of dual quaternions}
\label{sec fun dq}

Suppose we are given two unit dual quaternions, $\eta_1$ and $\eta_2$?  The problem is to find the dual quaternion valued function of time, denoted $\text{slerp}(\eta_1, \eta_2, t)$, such that at $t = 0$, it takes the value $\eta_1$, that at $t = 1$, it takes the value $\eta_2$, and such that its twist, $\theta$, is a constant vector dual quaternion.  The name of the function, `slerp', comes from the equivalent problem for orientations \cite{shoemake}.

This requires two functions.  The first is the exponential function:
\begin{equation}
\label{exponential}
\exp(\theta) = \sum_{k=0}^\infty \frac{\theta^k}{k!},
\end{equation}
The function $\exp(t \theta)$ as function of time $t$ is the pose traveled if it is the identity pose at $t = 0$, and maintains a constant twist $\theta$, that is
\begin{equation}
\frac d{dt} \exp(t\theta) = \theta \exp(t\theta) = \exp(t\theta) \theta .
\end{equation}
The second is a logarithm function, which given a unit dual quaternion $\eta$, finds $\theta$ such that $\exp(\theta) = \eta$.  There are infinitely many possibilities for $\theta$, so we will just describe the principal value, that is, the one for which the norm of the angular velocity is minimized.  Then
\begin{equation}
\begin{aligned}
\text{slerp}(\eta_1, \eta_2, t) &= \eta_1 \exp(t \log(\eta_1^* \eta_2)) \\
&= \exp(t \log(\eta_2 \eta_1^*)) \eta_1 .
\end{aligned}
\end{equation}
If one is merely interested in interpolating between the poses represented by these unit dual quaternions, multiply one of them by $\pm 1$ so that non-dual part of $\eta_1 \eta_2^{-1}$ has non-negative real part.

The results of this section may be found in \cite{montgomery-smith}.  We have seen similar results in \cite{selig,wang-et-al,wu-et-al}, but we believe our formulas to be more explicit.

Given any dual quaternion
\begin{equation}
\eta = c + \bm d + \epsilon (x + \bm y),
\end{equation}
by decomposing $\bm y$ into a vector parallel to $\bm d$ and perpendicular to $\bm d$, we can suppose without loss of generality that there exists orthonormal vectors $\bm a$ and $\bm b$ such that $\eta$ is a linear combination of $1$, $\bm a$, $\epsilon$, $\epsilon \bm a$, and $\epsilon \bm b$.

If $\bm a$ and $\bm b$ are orthonormal vectors, then if $w \ne 0$
\label{exp theta}
\begin{equation}
\begin{aligned}
& \exp\left(\tfrac12wt \bm a + \epsilon(\tfrac12v_1t \bm a + \tfrac12v_2t \bm b)\right) \\
&= \left(\cos(\tfrac12wt) + \sin(\tfrac12wt) \bm a\right) \left(1 + \tfrac12 \epsilon v_1t \bm a \right) \\
&\phantom{{}={}} + \epsilon \frac{v_2}w\sin(\tfrac12 wt) \bm b ,
\end{aligned}
\end{equation}
and if $w = 0$
\begin{equation}
\exp\left(\tfrac12 \epsilon vt \bm a\right) = 1 + \tfrac12 \epsilon v t \bm a .
\end{equation}
If $w \ne 0$, this represents a pose that rotates counterclockwise with angular velocity $w$ around $\bm a$, and translates by
\begin{equation}
\bm t = v_1 t \bm a + \frac {v_2}w\bigl(\sin(wt)  \bm b + (1-\cos(wt)) (\bm a \times \bm b) \bigr) ,
\end{equation}
that is, the sum of a point on a line in the direction of $\bm a$ moving with speed $v_1$, and a point on a circle in the plane perpendicular to $\bm a$ of radius $v_2/w$ which is traversed once every $2\pi/w$ time units.  If $w = 0$, this represents a pose with the identity rotation and translating in a straight line with velocity $v$.

Suppose $\bm a$ and $\bm b$ are orthonormal vectors, and that
\begin{equation}
\label{eta for log}
\eta = c + s \bm a + \epsilon (x + y_1 \bm a + y_2 \bm b)
\end{equation}
is a unit dual quaternion (so $c^2 + s^2 = 1$ and $cx + sy_1 = 0$), with $c \ne -1$.  Let
\begin{equation}
t = \text{\rm atan2}(s,c),
\end{equation}
that is, the angle part of the polar coordinates of $(c,s)$, with $-\pi < t \le \pi$.  If $t \ne \pi$ then
\begin{equation}
\log(\eta) = t \bm a + \epsilon (c y_1 - s x ) \bm a
+ \epsilon \frac{t y_2}{s} \bm b,
\end{equation}
where if $s = 0$ we assume $y_2 = 0$, and we set $y_2/s = 0$.

It is not possible to assign a canonical principal value for $\log(-1)$, because it could be $\pi \bm a$ for any unit vector $\bm a$.

\section{Proofs}
\label{sec proofs}

\begin{proof}[Proof of Equations~\eqref{norm mult} and~\eqref{normalize mult}]
First note that any dual number commutes with any dual quaternion.  Thus
\begin{equation}
|\eta_1 \eta_2|^2 = \eta_2^* \eta_1^* \eta_1 \eta_2 = \eta_2^* |\eta_1|^2 \eta_2 = |\eta_1|^2 \eta_2^* \eta_2 = (|\eta_1| |\eta_2|)^2 ,
\end{equation}
and
\begin{equation}
\widehat{\eta_1 \eta_2}
= |\eta_1 \eta_2|^{-1} \eta_1 \eta_2
= |\eta_1|^{-1} | \eta_2|^{-1} \eta_1 \eta_2
= |\eta_1|^{-1} \eta_1 | \eta_2|^{-1} \eta_2
= \widehat\eta_1 \widehat\eta_2.
\end{equation}

\end{proof}

\begin{proof}[Proof of Equation~\eqref{twist as dual quaternion}]
First, if we differentiate $\eta^* \eta = 1$, we obtain $\varphi + \varphi^* = 0$, that is, $\varphi$ is vector.

Suppose the position $\bm s$ is the image of the constant position $\bm r$ under the pose represented by $(\mathsf R, \bm t)$ and also represented by $\eta = Q + \epsilon B$.  Differentiating equation~\eqref{pose action} we obtain
\begin{equation}
\label{dot s 1}
\dot{\bm s} = \mathsf R \starop{\bm w} \bm r + \mathsf R \bm v = \mathsf R(\bm w \times \bm r + \bm v ).
\end{equation}
Differentiating equation~\eqref{defn of pose on 3-vector} we obtain
\begin{equation}
\epsilon \dot{\bm s}
= \frac d{dt}\bigl(\eta(1+\epsilon \bm r)\overline\eta^*\bigr)
= \eta\bigl(\varphi(1+\epsilon \bm r) + (1+\epsilon\bm r)\overline\varphi^*\bigr)\overline \eta^* .
\end{equation}
After some manipulation, and setting $\varphi = \tfrac12\bm a + \tfrac12\epsilon \bm b$, we obtain
\begin{equation}
\label{dot s 2}
\dot{\bm s} = Q(\bm a \times \bm r + \bm b)Q^* .
\end{equation}
Comparing equations~\eqref{dot s 1} and~\eqref{dot s 2}, we obtain
\begin{equation}
\bm w \times \bm r + \bm v = \bm a \times \bm r + \bm b,
\end{equation}
and since this holds for all $\bm r$, the result follows.
\end{proof}

\section{Projection from matrices to orthogonal matrices}
\label{sec final note}

What would be a good projection of $(3 \times 3)$ matrices onto rotation matrices?  One suggestion is the solution to the Orthogonal Procrustes problem \cite{higham,wiki1}, that is, to use $\mathsf A \mapsto \widehat{\mathsf A} = (\mathsf A\mathsf A^T)^{-1/2} \mathsf A = \mathsf A (\mathsf A^T \mathsf A)^{-1/2}$, which works if we have $\det(\mathsf A) > 0$.  One way to compute it is via the singular value decomposition $\mathsf A = \mathsf U \mathsf \Sigma \mathsf V^T$, when $\widehat{\mathsf A} = \mathsf U\mathsf V^T$.  However this is computationally expensive.


\begin{thebibliography}{99}

\bibitem{adorno} Bruno Vilhena Adorno, Robot Kinematic Modeling and Control Based on Dual Quaternion Algebra -- Part I: Fundamentals, 2017, hal-01478225.

\bibitem{agrawal} Om Prakash Agrawal. Hamilton operators and dual-number-quaternions in spatial kinematics. Mechanism and Machine Theory, 22(6):569-575, Jan 1987.

\bibitem{quaternions1} S.L. Altmann, Hamilton, Rodrigues, and the Quaternion Scandal, Mathematics Magazine, Vol. 62, No. 5. (Dec., 1989), pp. 291-308.

\bibitem{ball} R.S. Ball, The theory of screws: A study in the dynamics of a rigid body. Hodges, Foster and Co., Dublin, 1876.

\bibitem{bottema-et-al} O. Bottema, B. Roth, Theoretical Kinematics, North-Holland, Amsterdam, 1979.

\bibitem{chirikjian-et-al} Gregory S Chirikjian, Robert Mahony, Sipu Ruan, and Jochen Trumpf. Pose Changes From a Different Point of View. Journal of Mechanisms and Robotics, 10(2):021008-021008-12, Feb 2018.

\bibitem{clifford} M.A. Clifford, Preliminary Sketch of Biquaternions, Proceedings of the London Mathematical Society, Volume s1-4, Issue 1, November 1871, Pages 381-395, \url{https://doi.org/10.1112/plms/s1-4.1.381}.

\bibitem{dooley-et-al} J. R. Dooley and J. M. McCarthy, Spatial rigid body dynamics using dual quaternion components, Proceedings. 1991 IEEE International Conference on Robotics and Automation, Sacramento, CA, USA, 1991, pp. 90-95 vol.1, doi: 10.1109/ROBOT.1991.131559.

\bibitem{gallardo-alvarado} Jaime Gallardo-Alvarado, Kinematic Analysis of Parallel Manipulators by Algebraic Screw Theory, Springer, Switzerland, 2016.

\bibitem{han-et-al} Da-Peng Han, Qing Wei, and Ze-Xiang Li, Kinematic Control of Free Rigid Bodies Using Dual Quaternions, International Journal of Automation and Computing
05(3), July 2008, 319-324, DOI: 10.1007/s11633-008-0319-1.

\bibitem{higham} N.J. Higham, Matrix nearness problems and applications, in M.J.C. Gover and S. Barnett, editors, Applications of Matrix Theory, pages 1-27. Oxford University Press, 1989.

\bibitem{kavan-et-al} L. Kavan, S. Collins, J. \u Z\'ara, C. O'Sullivan, Skinning with Dual Quaternions, \url{https://dl.acm.org/doi/pdf/10.1145/1230100.1230107}.

\bibitem{kavan-et-al-2} L. Kavan, S. Collins, J. \u Z\'ara, C. O'Sullivan, Geometric Skinning with Approximate Dual Quaternion Blending, ACM Transactions on Graphics, Vol. 27, No. 4, Article 105, Publication date: October 2008.

\bibitem{kenwright} Ben Kenwright, A Beginners Guide to Dual-Quaternions, What They Are, How They Work, and How to Use Them for 3D Character Hierarchies, \url{https://cs.gmu.edu/~jmlien/teaching/cs451/uploads/Main/dual-quaternion.pdf}.

\bibitem{kussaba-et-al} Hugo T.M. Kussaba, Luis F.C. Figueredo, Jo\~ao Y. Ishihara, and Bruno V. Adorno, Hybrid kinematic control for rigid body pose stabilization using dual quaternions, Journal of the Franklin Institute, 354(7):2769-2787, May 2017.

\bibitem{montgomery-smith} Stephen Montgomery-Smith, Functional calculus for dual quaternions. Adv. Appl. Clifford Algebras 33, 36 (2023). https://doi.org/10.1007/s00006-023-01282-y.

\bibitem{perez-et-al} A. Perez and J.M. McCarthy, Dual Quaternion Synthesis of Constrained Robotic Systems, ASME. J. Mech. Des. May 2004 126(3) 425-435.

\bibitem{quaternions2} J. Pujol, Hamilton, Rodrigues, Gauss, Quaternions, and Rotations: a Historical Reassessment, Commun. Math. Anal., Volume 13, Number 2 (2012), 1-14.

\bibitem{schilling1}  M. Schilling, Universally manipulable body models---dual quaternion representations in layered and dynamic MMCs, Auton Robot 30, 399 (2011), \url{https://doi.org/10.1007/s10514-011-9226-3 https://link.springer.com/article/10.1007/s10514-011-9226-3}.

\bibitem{schilling2} M. Schilling, Hierarchical Dual Quaternion-Based Recurrent Neural Network as a Flexible Internal Body Model, 2019 International Joint Conference on Neural Networks (IJCNN), 2019, pp. 1-8, doi: 10.1109/IJCNN.2019.8852328, \url{https://ieeexplore.ieee.org/abstract/document/8852328}.

\bibitem{selig-book} J M Selig, Geometric fundamentals of robotics, Springer-Verlag New York Inc., 2nd edition, 2005.

\bibitem{selig} J.M. Selig, Exponential and Cayley Maps for Dual Quaternions, Advances in Applied Clifford Algebras, 20(3-4):923-936, May 2010.

\bibitem{shoemake} Ken Shoemake, Animating Rotation with Quaternion Curves, SIGGRAPH 1985, \url{https://www.cs.cmu.edu/~kiranb/animation/p245-shoemake.pdf}.

\bibitem{silva-et-al} Frederico F. A. Silva, Juan Jos\'e Quiroz-Oma\~na, and Bruno V. Adorno, Dynamics of Mobile Manipulators using Dual Quaternion Algebra, Journal of Mechanisms and Robotics, pages 1-24, Apr 2022.

\bibitem{spong-et-al} Mark W. Spong, Seth Hutchinson, and M. Vidyasagar, Robot Modeling and Control, Wiley, 2006.

\bibitem{ros} Stanford Artificial Intelligence Laboratory et al., Robotic Operating System, \url{https://www.ros.org}, 2018.

\bibitem{wang-et-al} Xiangke Wang, Dapeng Han, Changbin Yu, and Zhiqiang Zheng, The geometric structure of unit dual quaternions with application in kinematic control, Journal of Mathematical Analysis and Applications 389(2), 2012, 1352-1364.

\bibitem{wu-et-al} Yuanqing Wu, J.M. Selig and Marco Carricato,  (2019) Parallel
Robots with Homokinetic Joints: The Zero-Torsion Case, In: Uhl T. (eds) ``Advances in Mechanism and Machine Science.'' IFToMM WC 2019. Mechanisms and Machine Science, vol 73. Springer, Cham, pp. 269-278.

\bibitem{wiki0} Wikipedia, Hodge star operator, \url{https://en.wikipedia.org/wiki/Hodge_star_operator#Three_dimensions}.

\bibitem{wiki1} Wikipedia, Orthogonal Procrustes problem, \url{https://en.wikipedia.org/wiki/Orthogonal_Procrustes_problem}.

\bibitem{yang-et-al} XiaoLong Yang, HongTao Wu, Yao Li, Bai Chen, A dual quaternion solution to the forward kinematics of a class of six-DOF parallel robots with full or reductant actuation, Mechanism and Machine Theory 107 (2017) 27-36, \url{http://dx.doi.org/10.1016/j.mechmachtheory.2016.08.003}.

\end{thebibliography}
\end{document}